\pdfoutput=1

\documentclass[11pt]{article}

\usepackage[preprint]{acl}

\usepackage{times}
\usepackage{latexsym}

\usepackage[T1]{fontenc}

\usepackage[utf8]{inputenc}

\usepackage{microtype}

\usepackage{inconsolata}

\usepackage{graphicx}

\usepackage{hyperref}       
\usepackage{url}            
\usepackage{booktabs}       
\usepackage{amsfonts}       
\usepackage{nicefrac}       
\usepackage{microtype}      
\usepackage{xcolor}         

\usepackage{authblk}
\usepackage{multirow}
\usepackage{amssymb}
\usepackage{dsfont}
\usepackage{amsmath}
\usepackage{enumitem}
\usepackage{natbib}
\usepackage{tabularx}
\usepackage{tcolorbox}
\usepackage{caption}        

\usepackage{enumitem}
\setlist[itemize,1]{left=0pt}

\title{Advancing Process Verification for Large Language Models via Tree-Based Preference Learning}

\author{
 \textbf{Mingqian He}\textsuperscript{1},
 \textbf{Yongliang Shen}\textsuperscript{1\textdagger},
 \textbf{Wenqi Zhang}\textsuperscript{1},
 \textbf{Zeqi Tan}\textsuperscript{1},
 \textbf{Weiming Lu}\textsuperscript{1}\textsuperscript{\textdagger}
\\
 \textsuperscript{1}Zhejiang University
\\
\texttt{\{mingqianhe, syl, zhangwenqi, zqtan, luwm\}@zju.edu.cn}
}

\begin{document}
\maketitle

\renewcommand{\thefootnote}{\textdagger}
\footnotetext{Corresponding author.}

\begin{abstract}

Large Language Models (LLMs) have demonstrated remarkable potential in handling complex reasoning tasks by generating step-by-step rationales. Some methods have proven effective in boosting accuracy by introducing extra verifiers to assess these paths.
However, existing verifiers, typically trained on binary-labeled reasoning paths, fail to fully utilize the relative merits of intermediate steps, thereby limiting the effectiveness of the feedback provided. To overcome this limitation, we propose \textbf{Tree}-based \textbf{P}reference \textbf{L}earning \textbf{V}erifier (Tree-PLV), a novel approach that constructs reasoning trees via a best-first search algorithm and collects step-level paired data for preference training.
Compared to traditional binary classification, step-level preferences more finely capture the nuances between reasoning steps, allowing for a more precise evaluation of the complete reasoning path.
We empirically evaluate Tree-PLV across a range of arithmetic and commonsense reasoning tasks, where it significantly outperforms existing benchmarks. For instance, Tree-PLV achieved substantial performance gains over the Mistral-7B self-consistency baseline on GSM8K (67.55\% → 82.79\%), MATH (17.00\% → 26.80\%), CSQA (68.14\% → 72.97\%), and StrategyQA (82.86\% → 83.25\%).
Additionally, our study explores the appropriate granularity for applying preference learning, revealing that step-level guidance provides feedback that better aligns with the evaluation of the reasoning process.

\end{abstract}

\section{Introduction}
Large Language Models (LLMs) have demonstrated the ability to decompose complex questions into step-by-step problem-solving processes \citep{brown2020language, achiam2023gpt, liu2023evaluating, frieder2024mathematical}, achieving strong reasoning performance across a variety of tasks.

\begin{figure*}[ht]
\centering
\includegraphics[width=0.95\linewidth]{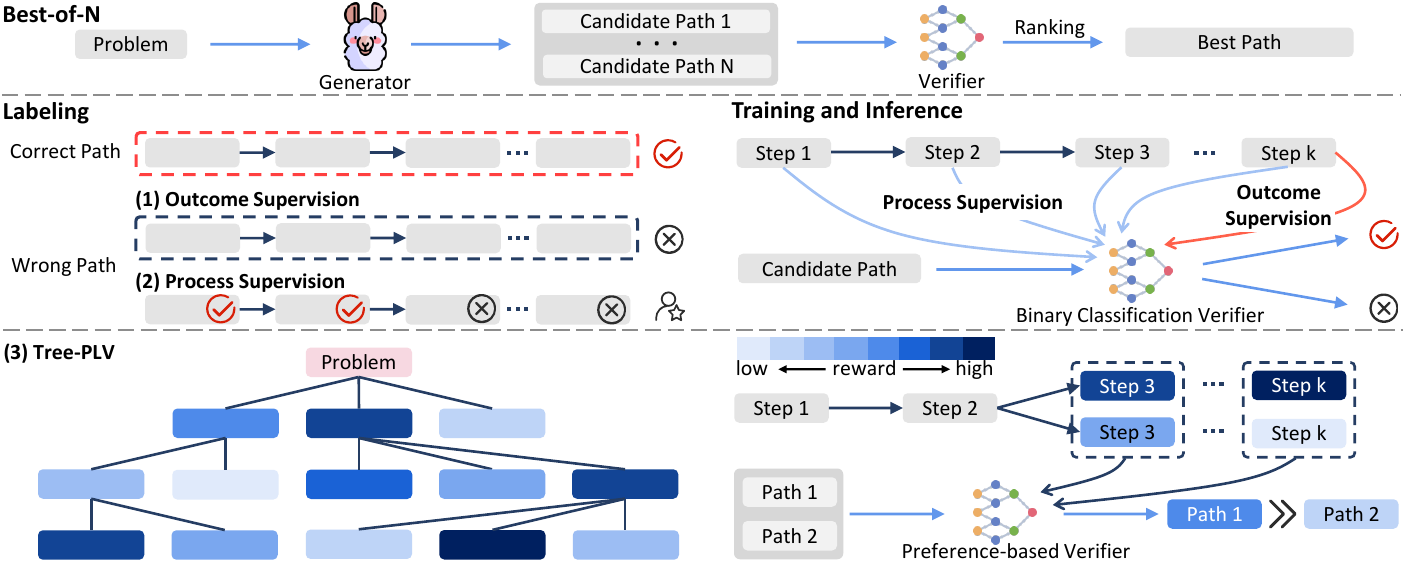}
\caption{A comparison of different methods: Traditional verifiers rely on binary labels for outcome and process supervision, whereas Tree-PLV employs preferences instead of scalar values.}
\label{fig:new_intro}
\end{figure*}

To enhance the reliability of reasoning paths, the best-of-N decoding strategy \citep{nakano2021webgpt, Askell2021AGL, cobbe2021training} is employed, where N candidate solutions are generated by the LLMs, and the most plausible one is selected based on specific rules \citep{golovneva2022roscoe, prasad2023receval}, such as coherence, logical consistency, and alignment with known facts.
Recently, some studies have introduced an auxiliary model, termed a verifier \citep{cobbe2021training}, to assess the quality of candidate solutions.
The training of verifiers can be divided into outcome supervision \citep{cobbe2021training,yu2023outcome,hosseini2024v} and process supervision \citep{Li2022MakingLM,lightman2023let,wang2023math} based on the granularity of the supervision signal (see Figure~\ref{fig:new_intro}).
Outcome supervision labels the entire path based on the final result, while process supervision evaluates the correctness of each individual step.

Whether employing outcome supervision or process supervision, verifiers are typically trained using binary classification \citep{uesato2022solving, lightman2023let, wang2023math}, which may not align ideally with the goals of optimizing verifiers. In the best-of-N decoding, verifiers are expected to rank candidate paths accurately. However, binary labels, offering rudimentary correct or incorrect signals, fail to capture the relative merits of different paths. Such coarse supervisory signals are insufficient to provide the detailed feedback necessary for verifiers to discern which steps are more effective, thereby limiting the potential for further improvements.
Moreover, the annotations used in training, often derived from answers, inherently contain some degree of noise. Even if the final answer is correct, the reasoning process may not be entirely accurate. Unfaithful reasoning and spurious shortcuts can also lead to the correct answer \citep{creswell2022faithful, lyu2023faithful, turpin2024language}. Consequently, training verifiers using binary classification is particularly vulnerable to noisy labels, which constrains the verifier’s capacity to precisely validate the steps.

To tackle these challenges, we propose a shift from a binary to a preference-based verifier. Trained through preference learning, it ranks the relative merits of different reasoning paths, allowing for more nuanced partitioning than simply judging them as correct or incorrect. 
The advantages of adopting this step-level preference-based verifier for ranking the reasoning paths include:

\begin{itemize}
    \item \textbf{Granular Validation at the Step Level:} Verifiers based on preference learning can capture subtle differences between steps, thereby providing more precise feedback.
    \item \textbf{Improving Verifier Robustness:} Focusing on ranking rather than binary judgments enhances the verifier's stability. As long as the relative ordering of steps is consistent, the training of the verifier remains robust against label noise.
    \item \textbf{Enhancing Model Explainability:} The detailed feedback provided by preference learning offers deeper insights into the reasoning process, moving beyond mere correctness on the final result.
\end{itemize}

Therefore, we introduce \textbf{Tree}-based \textbf{P}reference \textbf{L}earning \textbf{V}erifier (Tree-PLV), a novel method inspired by preference learning principles. Tree-PLV transcends traditional verifiers by modeling rewards based on comparisons between paths. Our method not only focuses on instance-level rewards derived from the outcomes but also emphasizes step-level optimization. This allows Tree-PLV to utilize intermediate steps to provide more finely grained feedback. Specifically, we employ a best-first search strategy during inference to construct a reasoning tree, with the initial problem statement as the root and each step as a node. Upon developing the tree, we construct our dataset by tracing paths from the root to each leaf node. At each level, we form pairs by conducting pairwise comparisons among child nodes, preferring those with higher rewards. This dataset serves to train our verifier using a ranking loss, greatly enhancing its ability to discern subtle nuances in reasoning sequences.

We conduct an empirical evaluation of Tree-PLV across diverse reasoning tasks, focusing on arithmetic reasoning with the GSM8K \cite{cobbe2021training} and MATH \cite{hendrycks2021measuring} datasets, and commonsense reasoning on the CSQA \cite{talmor2018commonsenseqa} and StrategyQA \cite{geva2021did} datasets. We benchmark Tree-PLV against existing verifiers, including self-consistency \cite{wang2022self} as a strong baseline. Our results indicate substantial performance gains across all datasets. For instance, when compared to the Mistral-7B self-consistency baseline, our method showed the following increases in accuracy: GSM8K (67.55\% → 82.79\%), MATH (17.00\% → 26.80\%), CSQA (68.14\% → 72.97\%), and StrategyQA (82.86\% → 83.25\%). Notably, Tree-PLV, when trained with data from GSM8K, demonstrates robust generalization to the more challenging MATH dataset.


\begin{figure*}[ht] 
\centering 
\includegraphics[width=0.95\linewidth]{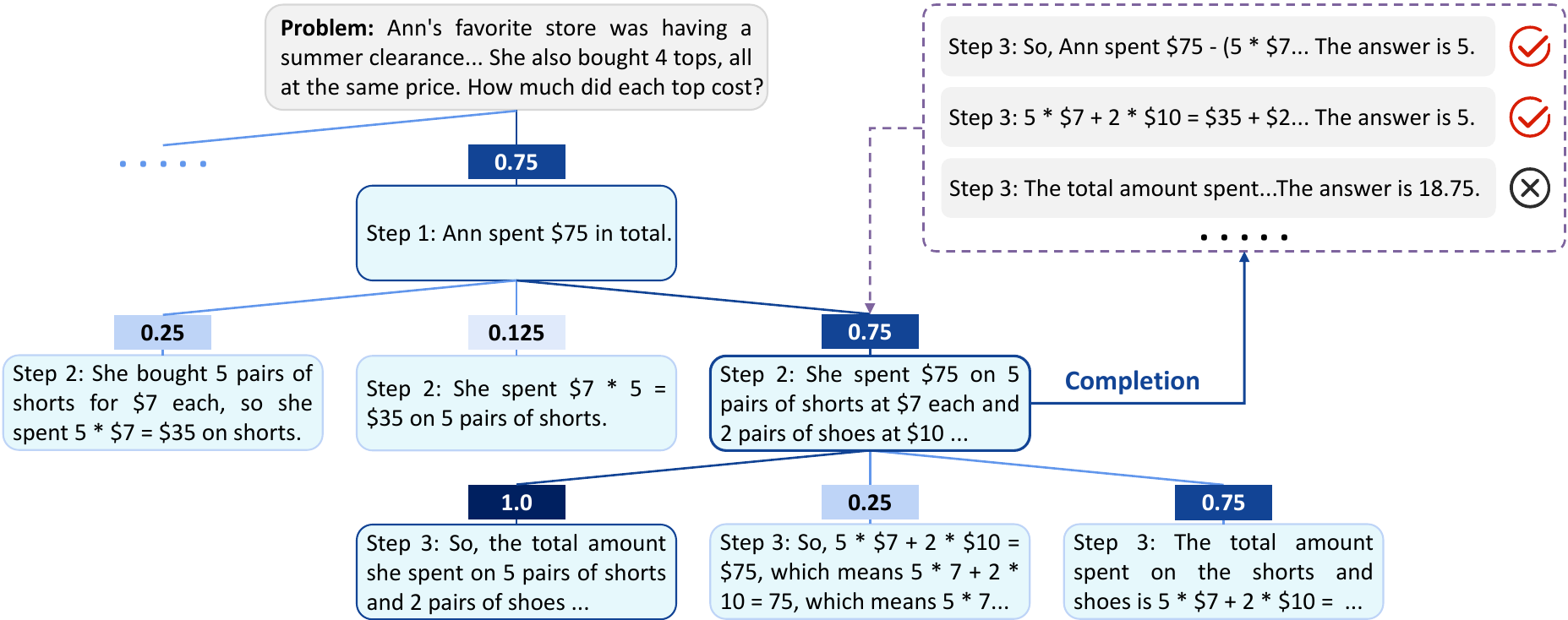}
\caption{The construction process of the reasoning tree. Best-first search consistently selects the child node with highest reward for further expansion. To evaluate the quality of the $i$-th step, we sample $N$ completions from it, denoted as $\mathcal{P}_i$. The reward is then calculated based on the proportion of these $N$ paths that yield the correct answer.} 
\label{fig:method} 
\end{figure*}

\section{Tree-PLV}
In this section, we introduce the Tree-PLV method, which leverages tree-based preference learning to advance verification for large language models in the context of stepwise reasoning processes. We begin by outlining the problem formulations (§~\ref{2.1}). Next, we detail how to construct a reasoning tree that represents reward preferences at each step (§~\ref{2.2}). Finally, we describe how we gather paired data for step-level preference learning and implement this into our verifier training (§~\ref{2.3}).

\subsection{Problem Formulations}
\label{2.1}
Following the best-of-N evaluation protocol proposed by \citet{lightman2023let}, we generate \(N\) candidate solutions \(\{y^{(1)}, y^{(2)}, \ldots, y^{(N)}\}\) from the generator model for a given input \(x\). Each solution \(y\) consists of a sequence of steps \(\{y_1, y_2, \ldots, y_n\}\). These solutions are then ranked by the verifier, and the highest-rated one is selected as the most plausible solution.

\subsection{Reasoning Tree Construction}
\label{2.2}
To provide precise step-level preference feedback, we implement a best-first tree search algorithm designed to generate paired data critical for preference learning. As Figure~\ref{fig:method} depicts, our method constructs a reasoning tree step-by-step, where each node represents a reasoning step. Expansion starts from the root of the tree at each search iteration.

At step $i$ of the tree expansion, we have a partial solution $y_{1:i-1}$ consisting of the previous $i-1$ reasoning steps. We use a reward function $\mathcal{R}(y_i | x, y_{1:i-1})$ to evaluate the quality of the next potential step $y_i$, given the input $x$ and the current partial solution $y_{1:i-1}$.
The tree search proceeds by expanding the most promising node at each iteration, i.e., the node whose child (the next potential step) has the highest reward according to $\mathcal{R}$. This guided exploration allows us to construct high-quality reasoning paths through the tree, providing paired data for preference learning between competing steps.

The traditional approaches regard the correctness of a step as its quality, relying on metrics like perplexity (PPL) or self-evaluation by LLMs to design the reward function $\mathcal{R}$. However, recent studies have shown that LLMs frequently struggle to effectively recognize errors \citep{huang2023large, hong2023closer, ren2023investigating}, which can degrade performance. To address this, we leverage the model's look-ahead capability to assess a step's quality by its potential to lead to the correct conclusion. Specifically, to evaluate a candidate step $y_i$, we use the same model to simulate $N$ subsequent reasoning trajectories starting from $y_i$, denoted as $N$ completions $\mathcal{P}_i = \{P_{i}^1, P_{i}^2, \ldots, P_{i}^N\}$. The quality of the step $y_i$ is quantified by the proportion of trajectories reaching the correct answer:
\begin{equation}
    \mathcal{R}(y_i) =  \frac{\sum_{j=1}^{N} \mathds{1}[a[P_{i}^{j}] = g]}{N}
\end{equation}
where \(a[P_{i}^{j}]\) is the outcome of the \(j\)-th trajectory \(P_{i}^{j}\) and \(g\) represents the golden answer. 

After determining the node with the highest reward value according to $\mathcal{R}(y_i | x, y_{1:i-1})$, we expand the tree by generating new child nodes. To achieve this, we sample $k$ potential subsequent reasoning steps $y_{i+1}^{j} \sim \pi_\theta(y_{i+1} | x, y_{1:i})$ for $j = 1, \ldots, k$, where $\pi_\theta$ is the language model used for reasoning. Each of these candidate steps $\{y_{i+1}^j\}_{j=1}^k$ becomes a new child node connected to the previously selected node. If the selected node represents the last step, indicating the end of the reasoning chain, we omit the expansion phase, and this iteration concludes. Guided by the reward function $\mathcal{R}$, this approach ensures a systematic exploration and expansion of reasoning paths in the search tree.

\subsection{Step-Level Pairwise Training}
\label{2.3}
A reasoning tree illustrates all potential reasoning paths, starting from the root and branching out to various leaf nodes. Our objective is to create a dataset $\mathcal{D}_{\mathrm{pair}}$ consisting of pairs that express preferences of reasoning paths. We generate this dataset by tracing each unique path from the root to the leaves of the tree. 
Within this dataset, each entry consists of a triplet in the form $\{(x, y^+, y^-)\}$, where $x$ denotes the initial problem statement, $y^+$ is the preferred reasoning sequence that leads to an accurate solution, and $y^-$ is a less desirable reasoning chain that results in an incorrect answer.

To collect the paired data $\{(x, y^+, y^-)\}$, we conduct pairwise comparisons between sibling nodes at each decision point along the tree. Sibling nodes are the various possible next steps in the reasoning process branching from the same prior context $y_{1:i-1}$.
If the reward difference between a preferable step $y_i^+$ and a less preferable step $y_i^-$ meets the minimum margin $\alpha$, we create a pair $(x, y^+, y^-)$ that includes the initial problem $x$, the superior reasoning sequence $y^+$, and the inferior alternative $y^-$. 
Specifically, for the preferable step \( y_i^+ \), we sample a path that leads to a correct solution among its completions \( \mathcal{P}_i^+ \). This constitutes the complete reasoning path \( y^+ \). Conversely, the less preferable step \( y_i^- \) is assembled with one of the subsequent paths \( \mathcal{P}_i^- \) that begins from it to form \( y^- \), which ultimately leads to incorrect outcomes.
Based on the pairwise instance data \( \{(x, y^+, y^-)\} \), we gather two types of step preferences during our training. The first type is derived from the reward function \( \mathcal{R} \) for the current step \( y_i \), which assesses the choices at each decision point. The second type of preference implicitly decomposes the outcome-guided instance preference into step-by-step comparisons, examining each potential path leading to possible outcomes. This breakdown allows us to allocate the instance preference to both the decision points and the completion steps in the reasoning process, ensuring that steps on the correct path are preferred over those on the incorrect path.



The verifier is built upon a large language model with an additional randomly initialized linear layer that outputs a scalar value. During evaluation, we take the value linked to the end token to represent the aggregate reasoning score.


We train the verifier using a step-level ranking loss function designed to promote the choice of the most effective solutions, defined as:
\begin{equation}
    \mathcal{L} = -\sum_{i=d}^{n} \log \sigma \left(r_{\phi}(x, y_{1:i}^{+}) - r_{\phi}(x, y_{1:i}^{-})\right)
\end{equation}
Here, $d$ is the index where the two reasoning paths diverge, and \( r_{\phi}(x, y_{1:i}) \) represents the output of our verifier for an input \( x \) and a sequence of steps \( \{y_1, \ldots, y_i\} \) with parameters \( \phi \).

\begin{table*}[ht]
\small
 \centering
 \begin{tabular}{l l c c c c}
  \toprule
  \multirow{2}{*}[-0.6ex]{\textbf{Models}} & \multirow{2}{*}[-0.6ex]{\textbf{Verifiers}} & \multicolumn{2}{c}{\textbf{Arithmetic}} & \multicolumn{2}{c}{\textbf{Commonsense}} \\
  \cmidrule(lr){3-4} \cmidrule(lr){5-6}
  & & GSM8K & MATH500 & CSQA & StrategyQA \\
  \midrule
  \multirow{5}{*}{LLaMA2-7B} & Self-Consistency     & 33.97 & 4.00 & 55.12 & \underline{76.82} \\
                           & ORM                  & 55.72 & 5.80 & 56.27 &  61.95 \\
                   & Self-Explore      & 51.40 & \underline{7.20} & 57.82 & 72.47 \\
                   & Math-Shepherd     & \underline{58.30} & 6.00 & \underline{58.23} & 70.78 \\
                   & Tree-PLV (Ours)   & \textbf{62.70} & \textbf{14.00} & \textbf{64.21} & \textbf{79.81} \\
                        
  \midrule
  \multirow{5}{*}{LLaMA2-13B} & Self-Consistency     & 54.51 & 3.40 & 64.62 & \underline{78.77} \\
                           & ORM                  & 66.34 & 9.40 & 66.34 & 64.74 \\
                   & Self-Explore    & 65.66 & \underline{9.60} & 66.67 & 78.64 \\
                   & Math-Shepherd     & \underline{68.01} & 6.60 & \underline{67.89} & 71.23 \\
                   & Tree-PLV (Ours)   & \textbf{76.12} & \textbf{18.20} & \textbf{71.66} & \textbf{83.64} \\
  \midrule
  \multirow{5}{*}{Mistral-7B} & Self-Consistency     & 67.55 &  17.00 & 68.14 &  \underline{82.86} \\
                            & ORM                  & 70.74 & 14.40 & 63.96 & 67.14 \\
                    & Self-Explore    & 72.18 & 20.00 & 67.49 & 77.99 \\
                    & Math-Shepherd     & \underline{74.91} & \underline{21.20} & \underline{70.11} & 79.22 \\
                    & Tree-PLV (Ours)   & \textbf{82.79} & \textbf{26.80} & \textbf{72.97} & \textbf{83.25}  \\
  \bottomrule
 \end{tabular}
 \caption{Results comparison (accuracy \%) on arithmetic and commonsense reasoning tasks, with various generators. \textbf{Bold} indicates the best results and \underline{underline} indicates the second best.}
 \label{tab:mainrst}
\end{table*}

\section{Experiments}
\subsection{Experimental Setup}
\paragraph{Tasks and Datasets}
In our evaluation, we select benchmarks from two distinct reasoning categories: arithmetic reasoning and commonsense reasoning. These tasks encompass diverse reasoning paradigms, enabling a comprehensive assessment of our method's effectiveness. 
For arithmetic reasoning, we utilize GSM8K \citep{cobbe2021training} and MATH \citep{hendrycks2021measuring}. GSM8K comprises grade school math problems, whereas MATH includes complex competition-level math problems. We use the entire GSM8K test set and a subset of 500 problems from MATH, which is identical to the test set of \citet{lightman2023let}. 
For commonsense reasoning, we employ CSQA \citep{talmor2018commonsenseqa} and StrategyQA \citep{geva2021did}. CSQA challenges the model with multiple-choice questions that often require reasoning based on complex semantics and prior knowledge. StrategyQA involves true-or-false questions that demand implicit multi-hop reasoning to derive answers.
\paragraph{Metrics}
Following the methodology suggested by \citet{lightman2023let}, we adopt a best-of-N evaluation paradigm. For arithmetic reasoning, we generate 64 solutions for each problem. However, for commonsense reasoning, due to the relatively limited search space for answers in these tasks, we generate 10 solutions for each question.
\paragraph{Models}
Our experiments leverage a series of models to serve as the generator, including LLaMA2-7B/13B \citep{touvron2023llama} and Mistral-7B \citep{jiang2023mistral}. For arithmetic reasoning tasks, we additionally conduct experiments with two specialized models: a version of Mistral-7B fine-tuned on MetaMATH \citep{yu2023metamath}, and WizardMath-7B \citep{luo2023wizardmath}.
\begin{table}[!ht]
    \centering
    \small
    \begin{tabular}{lcc}
        \toprule
        \textbf{Method} & \textbf{GSM8K} & \textbf{MATH500} \\
        \midrule
        Mistral-7B: MetaMATH \\
        \cmidrule{1-1}
        Self-Consistency & 83.55 & 35.00 \\
        ORM              & 85.67 & 29.20 \\
        Self-Explore     & 86.05 & 34.80 \\
        Math-Shepherd    & \underline{87.11} & \underline{35.40} \\
        Tree-PLV (Ours)  & \textbf{87.72} & \textbf{37.20} \\
        \midrule
        WizardMath-7B \\
        \cmidrule{1-1}
        Self-Consistency & 88.93 & 38.80 \\
        ORM              & 84.31 & 24.00 \\
        Self-Explore     & 88.86 & 39.60 \\
        Math-Shepherd    & \underline{89.16} & \underline{39.80} \\
        Tree-PLV (Ours)  & \textbf{90.14} & \textbf{40.20} \\
        \bottomrule
    \end{tabular}
    \caption{Results (accuracy \%) of the arithmetic reasoning task on generators with stronger capabilities.}
    \label{tab:additional}
\end{table}

\paragraph{Data Collection}
To construct the training dataset, we selected 6,000 problems from each of the GSM8K and CSQA training sets to generate paired data. We applied a margin threshold of $\alpha=0.375$ to filter these pairs, resulting in approximately 100k and 120k valid pairs for GSM8K and CSQA, respectively. For additional evaluation, we sampled 750 questions from the StrategyQA training set, which yielded 15k pairs. The verifier then trains for an epoch on the corresponding dataset based on task types. 
\renewcommand{\thefootnote}{2}
See Appendix~\ref{Experimental Details} for more detailed training parameter settings\footnote{The code is available at \url{https://github.com/Hareta-Leila/Tree-PLV}.}.

\paragraph{Baselines}
We conducted a comparative evaluation against several well-established verifiers. Specifically, we compared Tree-PLV with two verifiers trained using cross-entropy loss with binary labels: the outcome-supervised verifier, termed ORM \citep{lightman2023let}, and Math-Shepherd \citep{wang2023math}, a state-of-the-art process-supervised verifier that leverages automatically annotated data. Additionally, we adopted the self-consistency \citep{wang2022self} as a strong baseline, following \citet{lewkowycz2022solving}. Regarding paired data generation, we also considered Self-Explore \citep{hwang2024self} as a baseline method. It identifies the first erroneous step in a solution and uses the preceding steps to sample a correct path. These sequences are then paired to form training data. We evaluated the quality of data by comparing a verifier trained on this data with ours.

\begin{figure*}[!ht]
\centering
\includegraphics[width=\linewidth]{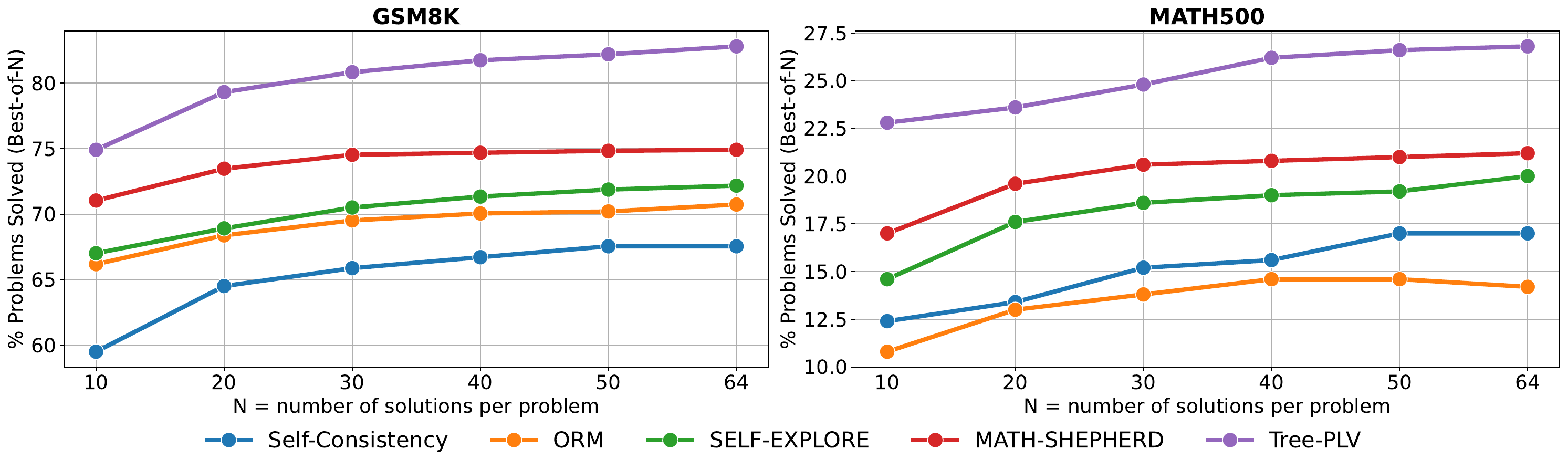}
\caption{Performance of different verifiers across varying numbers of solution (N) generated by Mistral-7B.}
\label{fig:modelperformance}
\end{figure*}

\subsection{Main Results}
\paragraph{Arithmetic Reasoning}
As shown in Tables~\ref{tab:mainrst}, our proposed method significantly outperforms other methods across all scenarios within the GSM8K and MATH500 datasets. For the GSM8K dataset, Tree-PLV consistently surpasses competing approaches.  For instance, using the LLaMA2-13B generator, Tree-PLV achieves an accuracy of 76.12\%, outperforming the second-best method, Math-Shepherd, by 8.11\%.
The Tree-PLV, initially trained on the GSM8K dataset, yields effective results on the more complicated MATH500 dataset, illustrating its strong generalization capabilities. Its adaptability stems from the training approach of Tree-PLV, which focuses on comparing the quality of steps rather than simply identifying correct or incorrect paths. This strategy makes the verifier more versatile.
Additionally, Tree-PLV requires only 22.7\% of the training data size used by Math-Shepherd, significantly reducing data requirements.

Table~\ref{tab:additional} displays the results of arithmetic reasoning tasks using advanced generators, namely Mistral-7B: MetaMATH and WizardMath-7B, on the GSM8K and MATH500 datasets. In all scenarios, our method, Tree-PLV, consistently outperforms the others. With Mistral-7B: MetaMATH, it achieves accuracies of 87.72\% on GSM8K and 37.20\% on MATH500. Similarly, with WizardMath-7B, Tree-PLV delivers the highest accuracies of 90.14\% and 40.20\% on GSM8K and MATH500, respectively. These results underline Tree-PLV's robust performance and adaptability in tackling complex arithmetic reasoning tasks across various generator settings.

\paragraph{Commonsense Reasoning}

The results in Table~\ref{tab:mainrst} highlights Tree-PLV's strong performance in commonsense reasoning tasks. Regardless of the dataset or the generator used, Tree-PLV consistently outperforms other verifiers, demonstrating its versatility and robustness. In the CSQA dataset, specifically, Tree-PLV improves accuracy by up to 5.98\%, 3.77\%, and 3.03\%, when compared with the second-best model. This high level of accuracy becomes even more pronounced with the StrategyQA dataset, confirming Tree-PLV's strong capabilities in handling intricate reasoning tasks. In contrast, both ORM and Math-Shepherd, which are trained using a binary supervision strategy, fail to show similar improvement. This underperformance could indicate the limitations inherent to binary supervision, especially in tasks that require complex reasoning. It is more prone to introducing noise into the binary supervision signal, which could consequently limit the verifier's capabilities.

We attribute these improvements to the following factors: 1) The training method of our verifier utilizes step-level preference learning, allowing for a nuanced evaluation of step quality that is better aligned with the best-of-N ranking paradigm. 2) The reward function, in our approach, is used to compare sibling steps instead of annotating them, which minimizes the impact of label noise. 3) Our approach improves the diversity of the data by incorporating comparisons among similar steps, thereby enriching the dataset and improving the robustness of our method.

\begin{figure*}[!ht]
\centering
\includegraphics[width=\linewidth]{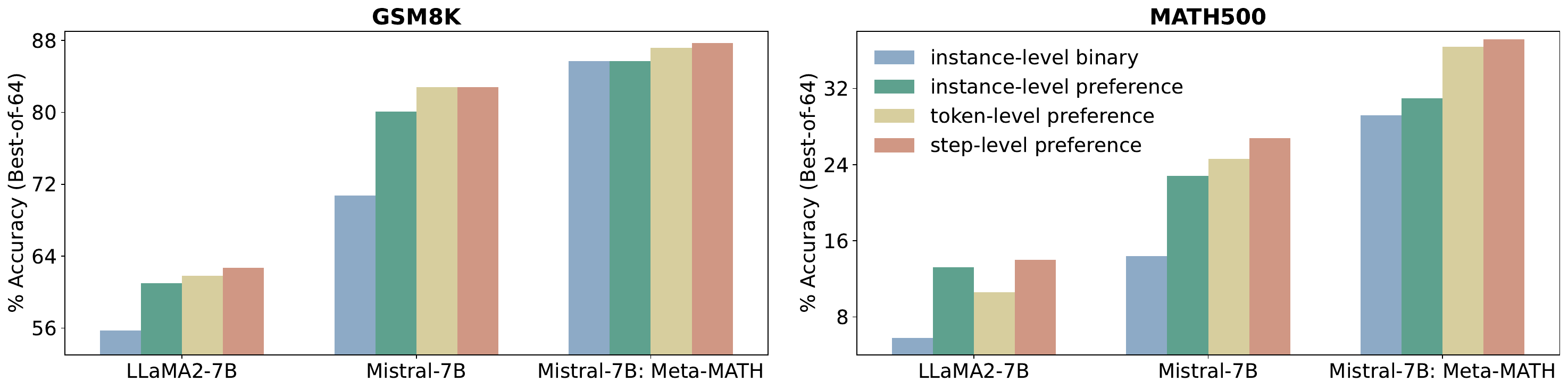}
\caption{A performance comparison of verifiers trained with different levels of feedback granularity.}
\label{fig:ablation}
\end{figure*}
\section{Analysis}
\subsection{Different Amounts of Candidate Solutions}
The number of candidate solutions impacts the performance of verification. As the number of solutions increases, more potentially superior candidates are introduced. However, there's a trade-off, as more solutions mean higher computational overhead and diminishing performance gains. For our study, we set a maximum of 64 solutions. As shown in Figure~\ref{fig:modelperformance}, we found that all methods show an enhancement in performance as the number of solutions increases and finally stabilizes at 64, consistent with \citet{wang2023math}'s findings. Throughout the study, the Tree-PLV method consistently outperforms the others across all quantities of solutions. Remarkably, as the number of solutions rises, Tree-PLV continues to widen the performance gap, underscoring its robustness and superior capacity to leverage a greater number of solutions.

\subsection{Granularity of Preference Learning}
We further analyze the appropriate granularity for applying preference learning, aiming to determine the most effective level of feedback for Tree-PLV. Specifically, we compare our step-level preference with instance-level and conventional token-level training methods \citep{christiano2017deep, ouyang2022training}. Additionally, we include a verifier trained using instance-level binary classification in our comparison for a comprehensive analysis. The results presented in Figure~\ref{fig:ablation} demonstrate that verifiers trained with preference learning consistently outperform those trained using binary classification across all levels of granularity. This finding suggests that preference learning is better aligned with the ranking evaluation pattern. Furthermore, step-level guidance exhibits the best performance, indicating that it represents the most suitable granularity. In contrast to instance-level sparse supervision, it provides more detailed and informative feedback. Moreover, in step-by-step inference scenarios, stepwise preference proves to be more reasonable than token-level supervision, and such step-level alignment is also more consistent with the intuitive human process of evaluating reasoning.

\subsection{Impact on Reasoning Path Distribution}
\begin{figure*}[!ht] 
\centering
\includegraphics[width=\linewidth]{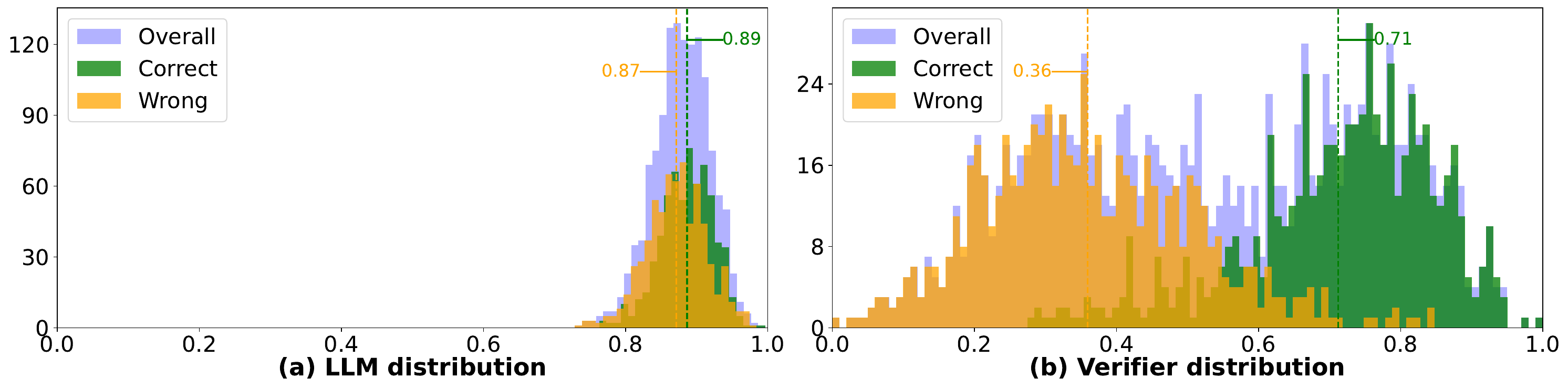}
\caption{Distributions of the LLM's generation confidence and verifier score on correct/incorrect reasoning paths. The median scores for correct and wrong paths are highlighted with lines of the same respective colors.}
\label{fig:distribution}
\end{figure*}

Recent studies have highlighted the difficulties large language models encounter in accurately identifying their errors \cite{huang2023large, hong2023closer}. To assess our verifier's effectiveness, we generate reasoning paths using greedy decoding, which are then scored by the original model's generation confidence and by Tree-PLV. We compare the score distributions from both methods. Figure~\ref{fig:distribution} shows Tree-PLV's capability to separate correct and incorrect reasoning paths. In contrast, generation confidence scores produce overlapping distributions for both categories, highlighting their limited differentiation capacity. Specifically, Tree-PLV increases the scores of correct paths by 35\%, whereas the confidence scores show only a marginal 2\% higher score for them. This pronounced disparity between the two methods not only confirms the superior performance of Tree-PLV in evaluating the quality of reasoning paths but also emphasizes the critical need for an effective verifier to accurately assess model outputs.

\begin{figure}[!ht] 
\centering
\includegraphics[width=\linewidth]{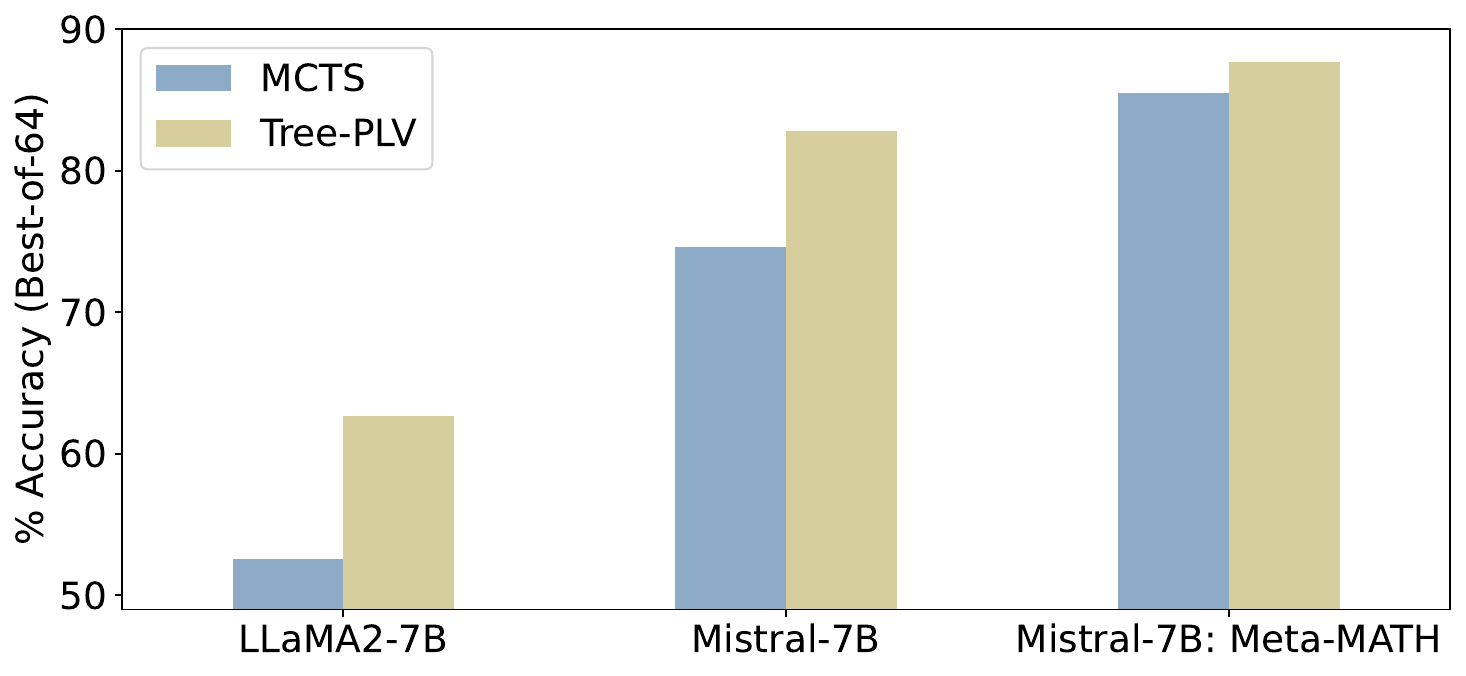}
\caption{Performance comparison of MCTS and Tree-PLV across different generators on GSM8K.}
\label{fig:mcts_1}
\end{figure}

\subsection{Reward Design}
\label{reward design}

Our proposed reward function $\mathcal{R}$ evaluates the quality of the current step by assessing the proportion of its latent complete paths that reach accurate conclusions. Some research demonstrates the advantages of integrating models' self-evaluation capabilities into inference to enhance accuracy \citep{hao-etal-2023-reasoning, ren2023self, Xie2023SelfEvaluationGB}. Inspired by this, we also developed a reward function that relies on self-evaluation. As per \citet{hao-etal-2023-reasoning}, we first construct a reasoning tree based on Monte Carlo Tree Search (MCTS) that incorporates self-evaluation. We then collect preference data from each step of this tree to train a verifier under the same setting. The results, as shown in Figure \ref{fig:mcts_1}, reveal that the self-evaluation-based reward does not perform as effectively as our proposed reward function. This trend is consistent across different generators, where the performance of the verifier trained with self-evaluation rewards significantly lagged behind our Tree-PLV. Our comparative analysis reveals that relying solely on the model's intrinsic ability to evaluate reasoning steps falls short in reliability. Additional results on the MATH500 dataset are provided in the Appendix~\ref{additional results}.




\begin{figure}[ht]
\centering 
\includegraphics[width=\linewidth]{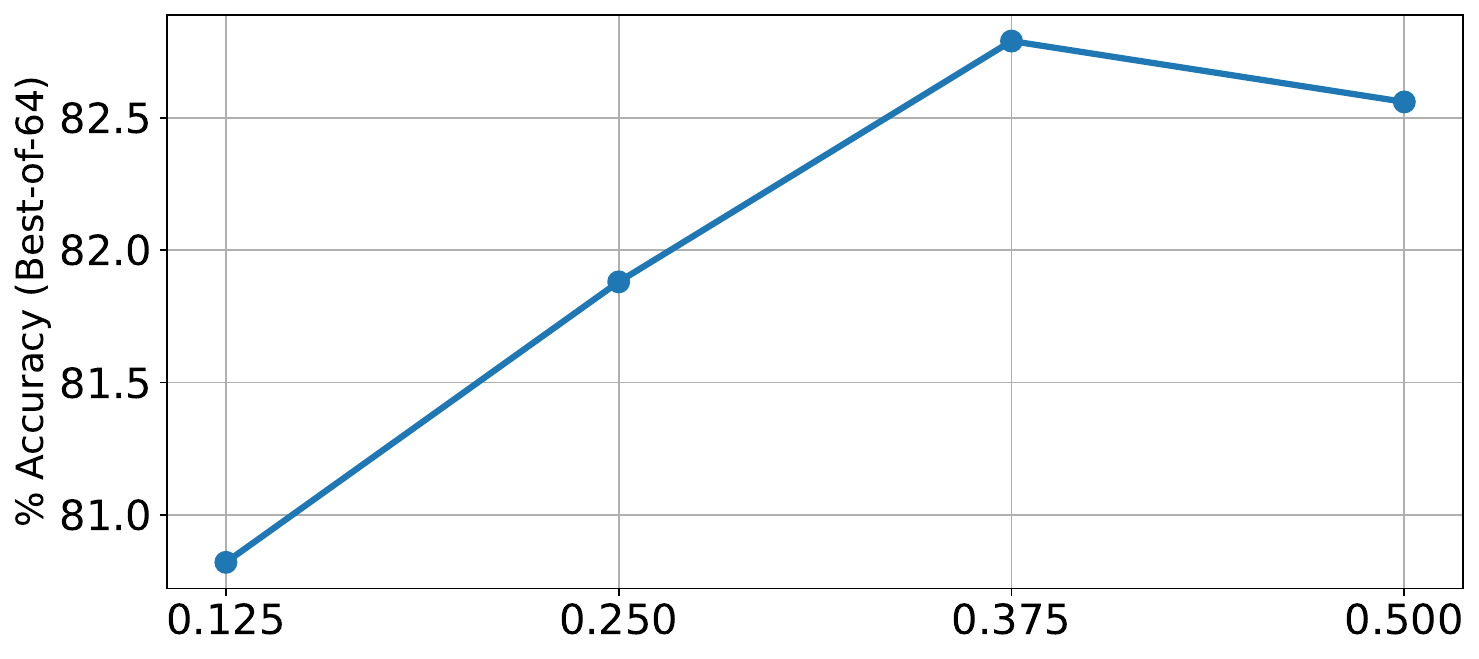}
\caption{Results on GSM8K, showing verifiers trained with data sampled using different margins.}
\label{fig:margin}
\end{figure}

\subsection{Margin Value for Preference Collection}
\label{margin}
We delve deeper into the analysis by applying varying margins to filter data. As shown in Figure~\ref{fig:margin}, increasing the margin value raises the proportion of paired data with more noticeable path differences, thereby reducing noise and improving the performance of the verifier. However, when the margin value reaches 0.5, the verifier's performance declines, indicating that excessively high margins can diminish the comparison of similar paths and reduce the richness of the data. Results on the MATH500 dataset are provided in the Appendix~\ref{additional results}.

\subsection{Efficiency of Data Generation}
\begin{figure}[ht]
    \centering
    \includegraphics[width=\linewidth]{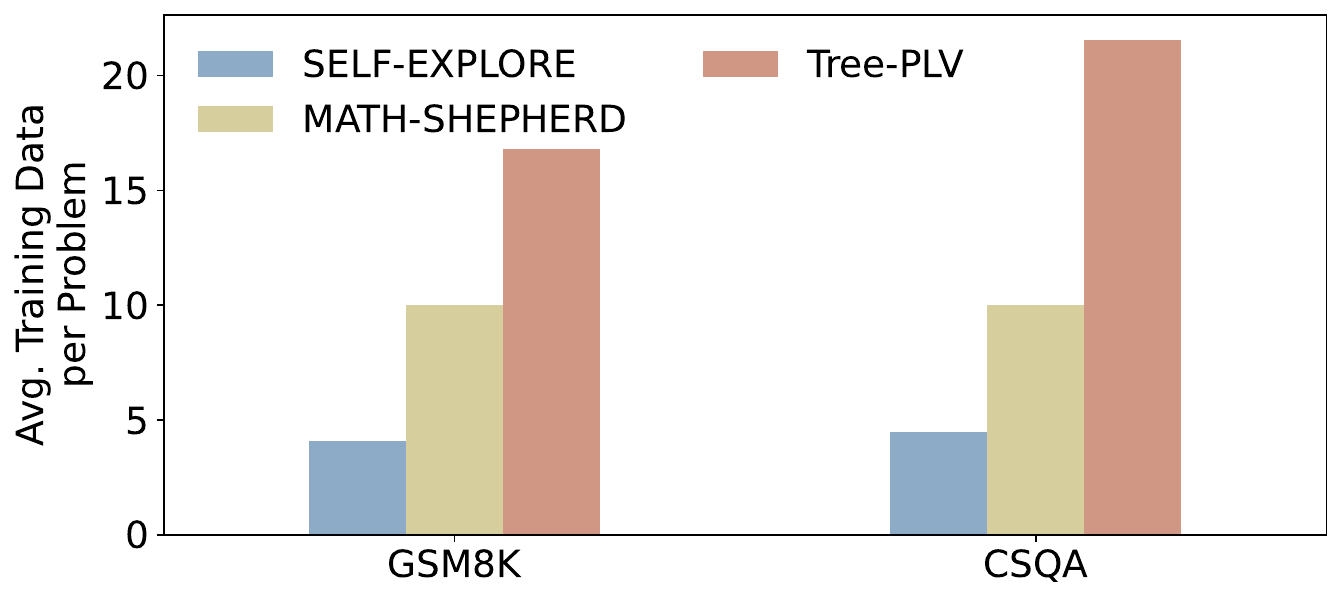}
    \caption{A comparison of methods' efficiency in generating training data.}
    \label{fig:numperiter}
\end{figure}

We further analyze our tree-based framework across the dimensions of data generation efficiency. We compare the average number of training data per problem generated by different methods over 10 iterations, as shown in Figure~\ref{fig:numperiter}. Math-Shepherd consistently labels ten paths. In contrast, Self-Explore struggles with datasets where the model already performs well, as it inherently relies on incorrect paths to form pairs. Our framework enhances the utilization of single paths through the branching structure of a tree, demonstrating significant scalability and efficiency in data generation.

\section{Related Work}
Recent studies \citep{lightman2023let, cobbe2021training, yu2023outcome} have demonstrated that incorporating a verifier can significantly enhance performance in reasoning tasks. Primarily, there are two methods for training verifiers: outcome supervision and process supervision. Unlike these approaches, which rely on supervised learning with binary labels, our methodology employs preference learning to achieve step-level alignment. We provide a detailed discussion of additional related work in Appendix~\ref{related work}.

\section{Conclusion}
In this paper, we propose Tree-PLV, an innovative verifier developed through step-level preference learning. Utilizing a best-first tree search framework to construct reasoning trees, this method creates a diverse and robust dataset, ensuring a rich pool of paired data for preference training.  By leveraging step-level preferences instead of conventional binary labels, our methodology provides more granular and detailed feedback, enhancing the verification of reasoning paths. 
Our empirical findings highlight the efficacy of Tree-PLV, showing significant performance improvements over existing benchmarks across a variety of arithmetic and commonsense reasoning tasks. In future work, we aim to delve deeper into integrating our verifier within the reasoning process by combining it with various search algorithms, thereby refining the verification of intermediate reasoning steps.

\section*{Limitations}
While Tree-PLV is effective at scoring complete solutions, its potential to provide feedback that assists and refines the reasoning process during the inference phase has not been explored in this work. Additionally, aside from assisting in reasoning to encourage successful outcomes, some studies treat verifiers as reward models in reinforcement learning scenarios. However, our current research focuses on developing highly reliable verifiers, rather than employing reinforcement learning to enhance the generator. Addressing these aspects will be the focus of our future research.

\section*{Ethics Statement}
The development of Tree-PLV aims to improve the accuracy and reliability of reasoning assessments in Large Language Models. Although our method poses no immediate ethical concerns, we acknowledge the potential for misuse if applied in sensitive areas such as automated decision-making. We recommend rigorous evaluation and oversight to prevent bias and ensure data privacy in all applications. It is crucial to maintain transparency and adhere to ethical standards in the deployment of such technologies.


\bibliography{acl_latex}

\clearpage
\appendix
\section{Related Work}
\label{related work}
\paragraph{Multi-step Reasoning}
Challenging reasoning tasks have spurred innovative research in large language models (LLMs), which are essential for handling complex queries \citep{kaddour2023challenges,lightman2023let, huang2023large}. A prominent strategy employed is the Chain-of-Thought (CoT) prompting technique \citep{wei2022chain}, along with its derivatives \citep{kojima2022large, wang2022self, yao2024tree}. These methods decompose the reasoning process into sequential steps, systematically approaching problem-solving by mimicking human-like reasoning. To further enhance the accuracy of these intermediate steps, recent studies leverage extensive synthetic datasets, which are either distilled from cutting-edge models \citep{yu2023metamath, luo2023wizardmath} or composed of self-generated rationales \citep{zelikman2022star, yuan2023scaling, ni2022learning}, to fine-tune the LLMs. Such training effectively sharpens the models' ability to produce CoT reasoning that leads to correct answers.

In addition, there is an increasing emphasis on test-time verification, a process that generates multiple solutions and ranks them through a separate verifier \citep{cobbe2021training} to select the most accurate one. The DIVERSE framework \citep{Li2022MakingLM} demonstrates the effectiveness of employing a variety of CoT prompts in conjunction with a verifier to tackle reasoning challenges. CoRe \citep{zhu2022solving} implements a dual-process system by fine-tuning both the generator and the verifier, thereby enhancing the performance of LLMs on math word problems.

\paragraph{Training Verifiers}
The training process for verifiers is typically approached as a binary classification task. In closely related work, \citet{uesato2022solving} introduces two distinct training methodologies for verifiers, differentiated by the granularity of the supervision signal: outcome supervision \citep{cobbe2021training,yu2023outcome,hosseini2024v} and process supervision \citep{Li2022MakingLM,uesato2022solving,lightman2023let,wang2023math}. Verifiers trained under outcome supervision utilize only the final result of a reasoning path for learning, while those trained under process supervision benefit from incremental feedback at each step within the CoT.

The training objective for this binary classification does not align well with its intended application in ranking scenarios. The binary labels provide a coarse supervisory signal, which fails to differentiate the efficacy of various steps, thus capping potential enhancements in performance. Addressing these issues is not straightforward, as accurately assessing the quality of each step involves subjective judgments that are difficult to standardize, even with human annotations. Math-Shepherd \citep{wang2023math} employs heuristic rules to annotate individual steps; however, the verifier trained on these refined labels does not show improvement compared to its binary version. Our method, Tree-PLV, fundamentally rethinks the verifier’s role by incorporating step-level preference learning. This approach more effectively captures subtle differences among paths and better aligns with the ranking evaluation paradigm.

\section{Experimental Details}
\label{Experimental Details}
\subsection{Dataset Description}
Table~\ref{tab:benckmark_example} presents the reasoning benchmarks used in this paper, along with examples. We utilize the same test sets as \citet{wang2023math} for GSM8K and MATH500, and as \citet{wei2022chain} for CSQA.

For StrategyQA, there are 2,290 test cases (i.e., questions paired with TRUE/FALSE labels). However, there are no additional cases that can be leveraged by Tree-PLC to construct the training set. To address this issue, we randomly sampled 750 test cases to create the training data, leaving the remaining 1,540 cases for testing.

\begin{table*}[!ht]
\centering
    \begin{tabularx}{\textwidth}{llX}
    \toprule
    \textbf{Dataset} & \textbf{N} & \textbf{Example Problem} \\
    \midrule
    GSM8K    & 1319 & James decides to run 3 sprints 3 times a week. He runs 60 meters each sprint. How many total meters does he run a week? \\
    \midrule
    MATH500  & 500 & A regular hexagon can be divided into six equilateral triangles. If the perimeter of one of the triangles is 21 inches, what is the perimeter, in inches, of the regular hexagon? \\
    \midrule
    CSQA    & 1221 & Where would you find magazines along side many other printed works? Answer Choices: (A) doctor (B) bookstore (C) market (D) train station (E) mortuary\\
    \midrule
    StrategyQA & 1540 & Could a silverfish reach the top of the Empire State Building? \\
    \bottomrule
    \end{tabularx}
    \caption{Reasoning benchmarks we use in this paper with examples. N: number of test cases.}
    \label{tab:benckmark_example}
\end{table*}

\subsection{Training Data}
To construct the training set for Tree-PLV, we select 6,000 problems each from the GSM8K and CSQA training datasets to generate paired data. We sample 8 subsequent trajectories (N=8) to calculate the reward for each step. By applying a margin threshold of \(\alpha=0.375\), we filter these pairs, resulting in approximately 100k valid pairs from GSM8K and 120k from CSQA. Additionally, we sampled 750 questions from the StrategyQA training set, yielding 15k pairs.


\subsection{Training Details}
Considering our limited computational resources, we use Mistral-7B as the backbone to train our verifier via the LoRA \citep{hu2021lora} training method for one epoch. We choose a learning rate of 1e-6, implemented with a cosine learning rate scheduler. Specifically for arithmetic reasoning, the base model is fine-tuned on MetaMATH, following \citet{wang2023math}. Furthermore, ORM, Self-Explore, and Math-Shepherd are trained under settings identical to those we use.
\subsection{Prompt Examples}
We present a prompt example for generating solutions and performing reasoning. We employ few-shot learning to guide the model towards producing outputs in the step-by-step format.
\begin{tcolorbox}[boxrule=0.5mm, left=1mm, right=1mm, top=1mm, bottom=1mm]
Solve the questions step by step. Make each step clear and logical, leading to the next. If a step directly leads to an answer, end it with "The answer is: ". If not, proceed to the next step until you find the solution.\\
\\
Four years ago, Kody was only half as old as Mohamed. If Mohamed is currently twice as 30 years old, how old is Kody?\\
Step 1: If Mohamed is currently twice as old as 30 years old, then Mohamed is currently 2 * 30 = 60 years old.\\
Step 2: Four years ago, Mohamed was 60 - 4 = 56 years old.\\
Step 3: Four years ago, Kody was only half as old as Mohamed, so Kody was (1/2) * 56 = 28 years old.\\
Step 4: Therefore, Kody is currently 28 + 4 = 32 years old. The answer is: 32.\\
\\
\textcolor{gray}{<more examples>}
\end{tcolorbox}

\subsection{Preference Pair}
We present an example of preference pairs $\{(x, y^+, y^-)\}$ used for training Tree-PLV, where $x$ represents the problem statement, $y^+$ denotes the preferred reasoning sequence, and $y^-$ is a less desirable path.

\begin{tcolorbox}[boxrule=0.5mm, left=1mm, right=1mm, top=1mm, bottom=1mm]
$x$: One pie costs \$4 for a piece. Each pie is having 3 pieces. During one hour the bakery can make 12 pies. Creating one pie costs the bakery \$0.5. Considering the bakery would be able to sell all pie pieces, how much money would it make?\\
\\
$y^+$: \\
Step 1: Each pie is having 3 pieces.\\
Step 2: 12 pies is equal to 3 * 12 = 36 pieces.\\
\textcolor{red}{Step 3: Each piece costs \$4, so the bakery would earn 36 * \$4 = \$144 if it could sell all of its pieces.}\\
Step 4: The cost of making one pie is \$0.5, so the cost of making 12 pies is 12 * \$0.5 = \$6.\\
Step 5: The bakery would make $144 - $6 = \$138. The answer is: 138.\\
\\
$y^-$: \\
Step 1: Each pie is having 3 pieces.\\
Step 2: 12 pies is equal to 3 * 12 = 36 pieces.\\
\textcolor{red}{Step 3: The bakery can make 36 pieces during one hour.}\\
Step 4: One piece costs \$4 and a pie costs \$0.5, so it costs the bakery (1/3) * \$4 + (2/3) * \$0.5 = \$1.5 to create one pie.\\
Step 5: The bakery would make 12 * \$1.5 = \$18 during one hour. The answer is: 18.
\end{tcolorbox}

\begin{figure*}[ht]
    \centering
    \begin{minipage}[b]{0.48\linewidth}
        \centering
        \includegraphics[width=\linewidth]{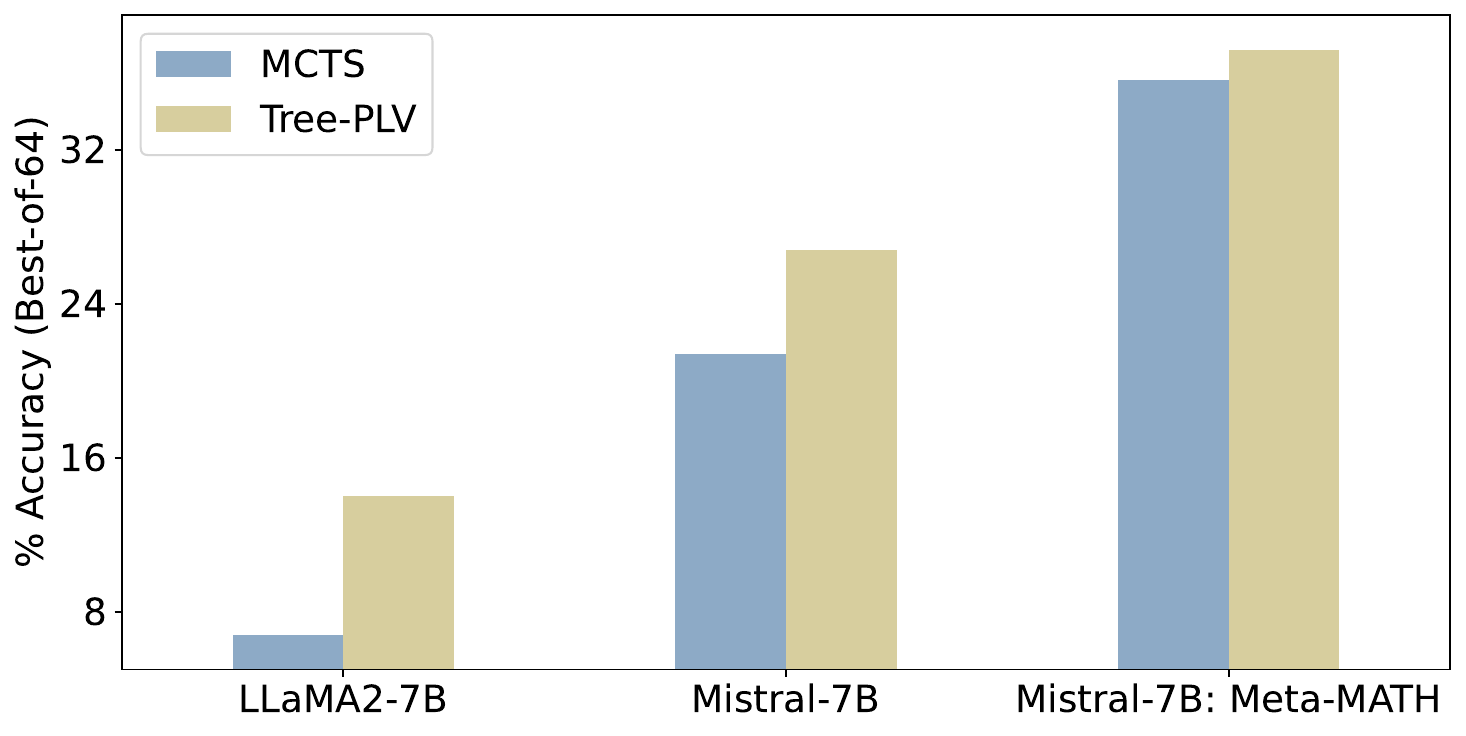}
        \caption{Performance comparison of MCTS and Tree-PLV across different generators on MATH500.}
        \label{fig:mcts_2}
    \end{minipage}
    \hfill
    \begin{minipage}[b]{0.48\linewidth}
        \centering
        \includegraphics[width=\linewidth]{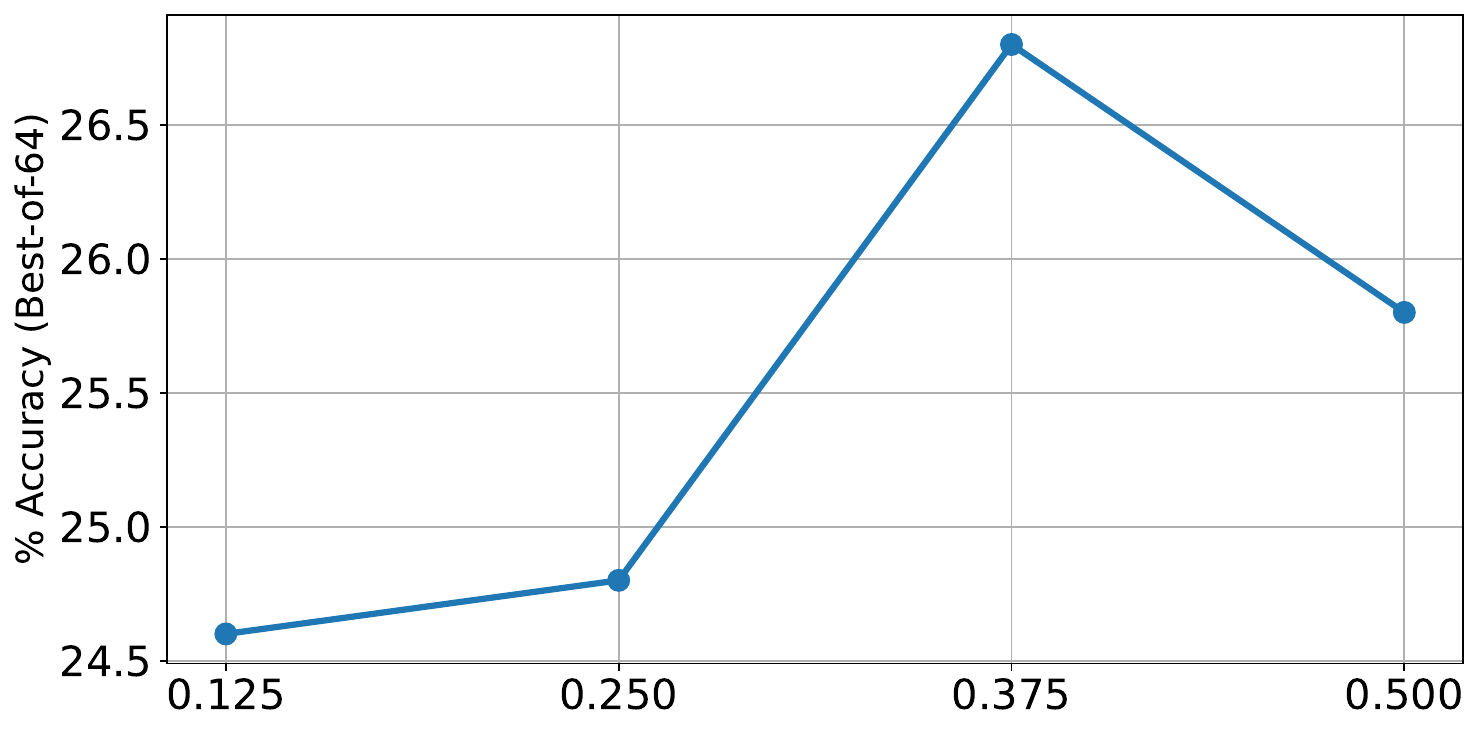}
        \caption{Results on MATH500, showing verifiers trained with data sampled using different margins.}
        \label{fig:margin_2}
    \end{minipage}
\end{figure*}

\section{Additional Experiment Results}
\label{additional results}
\subsection{Additional Results of Reward Design}
In Figure~\ref{fig:mcts_2}, we show the additional results on the MATH500 dataset for verifiers in Section~\ref{reward design}. The results of the experiment are consistent with our paper's analysis; our reward design is comparatively more effective.
\subsection{Additional Results on Different Margins}
In this section, we provide the experimental results of different verifiers on MATH500 in Section~\ref{margin}. As shown in Figure~\ref{fig:margin_2}, the performance trend of the verifiers with different margin values is consistent with the trend on GSM8K.

\end{document}